\documentclass[final]{cvpr}

\usepackage{times}
\usepackage{epsfig}
\usepackage{graphicx}
\usepackage{amsmath}
\usepackage{amssymb}
\usepackage{booktabs}
\usepackage{wrapfig}

\usepackage[pagebackref=true,breaklinks=true,colorlinks,bookmarks=false]{hyperref}

\def\cvprPaperID{2032} 
\def\confYear{CVPR 2021}

\begin{document}

\title{Tiny Video Networks}

\author{AJ Piergiovanni\\
Google Research\\
\and
Anelia Angelova \\
Google Research\\

\and
Michael Ryoo\\
Robotics at Google\\
}

\maketitle

\begin{abstract}
Automatic video understanding is becoming more important for applications where real-time performance is crucial and compute is limited: e.g., automated video tagging, robot perception, activity recognition for mobile devices. 
Yet, accurate solutions so far have been computationally intensive. 
We propose efficient models for videos - Tiny Video Networks - which are video architectures, automatically designed to comply with fast runtimes and, at the same time, are effective at video recognition tasks. The Tiny Video Networks run at faster-than-real-time speeds, demonstrating strong performance or outperforming state-of-the-art approaches across several video benchmarks.
Models are publicly available via TF.Hub at \href{https://tfhub.dev/google/tiny_video_net/mobile_1/1}{https://tfhub.dev/google/tiny$\_$video$\_$net/mobile$\_$1/1}
Code is available at \href{https://github.com/google-research/google-research/tree/master/tiny_video_nets}{https://github.com/google-research/google-research/tree/master/tiny$\_$video$\_$nets}.


\end{abstract}





\section{Introduction}
\label{intro}

\maketitle


Understanding videos is an important problem  in computer vision with many applications, such as automated video tagging, activity recognition, mobile and robot perception. At the same time video understanding is a challenging task as it needs to incorporate spatio-temporal information across multiple frames.  

Previous methods for video analysis use complex and computationally intensive models~\cite{tran2014c3d,carreira2017quo,xie2018rethinking,wang2018non}. 
Efficient video architectures have been recently designed~\cite{eco,lin2019tsm}. Architecture search has also been involved in producing efficient video networks, e.g. in X3D~\cite{feichtenhofer2020X3D}. 
These approaches are still not suitable for real-time video processing, which greatly hinders their application to real-world systems, e.g., mobile vision or robotics.

\begin{figure}
    \centering
     \includegraphics[width=0.8\linewidth]{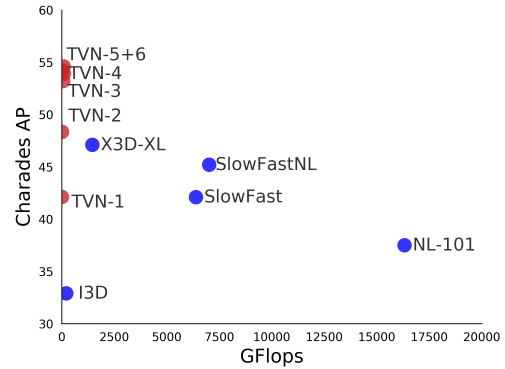}

    \caption{
    GFlops vs model accuracy of Tiny Video Networks (TVN) compared to the state-of-the-art.
    Tiny Video Networks are highly efficient video models surpassing previous powerful models X3D and SlowFast in both dimensions. Charades dataset.
    }
    \label{fig:intro}
    \vspace{-0.1cm}
\end{figure}

To address these challenges, we propose to automatically design video networks that provide strong recognition performance at a fraction of the computational cost with faster-than-real-time speeds.
More specifically, we propose a general video architecture search approach, based on evolution, which designs a family of `tiny' neural networks for video understanding (Figure~\ref{fig:intro}).
The networks achieve high accuracy and run efficiently, at real-time or better speeds, e.g. within 20 ms on a GPU per video clip (which is of $\sim$1 second duration and has 30 frames). 
We call them Tiny Video Networks (TVN), as they require extremely small runtimes, which is unprecedented for video models.
TVNs operate in the high accuracy and low runtime area of the accuracy-runtime curve where no other models exist (Figure~\ref{fig:intro}), and outperform state-of-the-art video models, despite being order, or orders, of magnitude faster. 

\begin{figure}
\includegraphics[width=\linewidth]{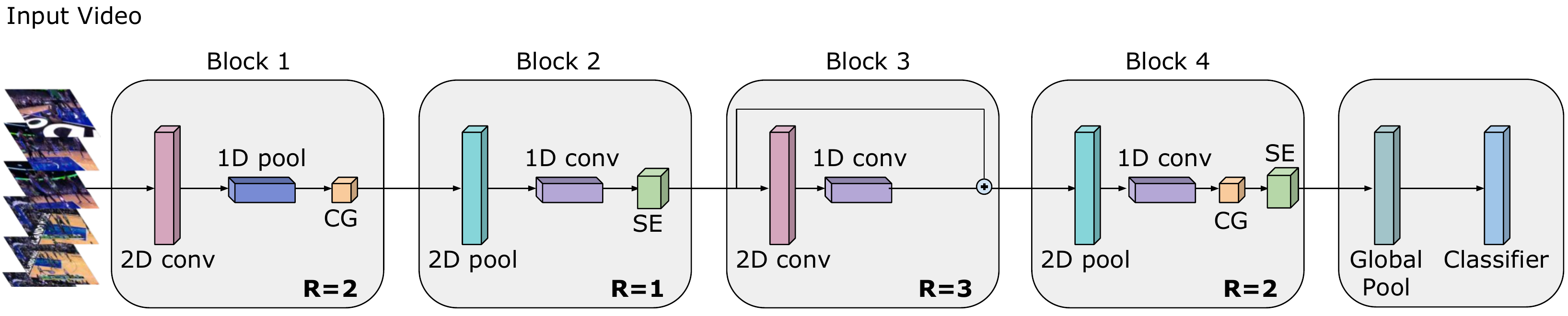}
   \includegraphics[width=\linewidth]{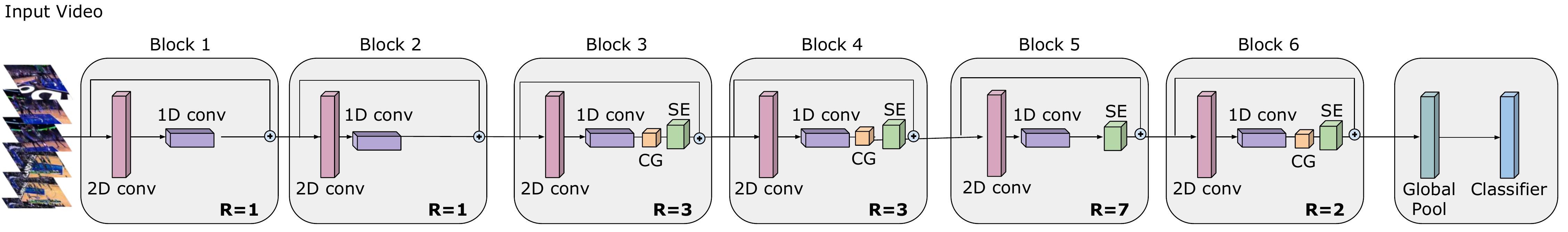}
  \caption{
  Example Tiny Video Networks, created from primitive elements using architecture search. TVN-1 (top) takes 37 ms on CPU (10ms on GPU) for inference. TVN-5 (bottom) takes 86ms on CPU and 16ms on GPU.
  Each net has multiple blocks, each repeated $R$ times. Each block has a different configuration with spatial and temporal convolution, pooling, non-local layers, context gating, etc. 
  The architecture search can also select the input resolution, the number of frames to sample and the frame rate.
}
  \label{fig:teaser}
  \vspace{-0.1cm}
\end{figure}

Architecture search for videos is a challenging task as the search needs to span the full spatio-temporal domain, while processing an much more data per example, e.g. 30x the frames. 
Neural architecture search (NAS) approaches~\cite{zoph2017neural,pham2018efficient,liu2019darts} have been  successful, 
but are time-intensive even for images.
In this work we address both problems, showing that it is possible to design successful video architectures via architecture search differently, both structurally and at the component level from low-level primitives, and that, at the same time, they can be very accurate and run at faster than real-time speeds. 

The contribution of this work is creating highly efficient and accurate Tiny Video Networks, with very strong performance on video understanding tasks, despite working at the fraction of the cost.  
We discover novel and interesting video architectures, which at the same time are easy to understand
(Figure~\ref{fig:teaser}). Furthermore, we propose a search space which enables the combination of efficient layers, which effectively interleave components working in the time and spatial domains. As a result, the architecture search is very efficient, despite being done on videos.
Thirdly, using the Tiny Video Networks architectures, we create TVNs suitable for mobile devices. Our Mobile TVNs significantly outperform the popular MobileNet models~\cite{mobilenetv2} applied to videos, both in accuracy and runtime. 

We evaluate TVNs on well-established video understanding tasks for multi-class and multi-label multi-class action classification and test their generalizability on four challenging datasets: 
Moments-in-Time \cite{monfort2018moments}, HMDB \cite{kuehne2011hmdb}, Charades \cite{sigurdsson2016hollywood} and MLB \cite{mlbyoutube2018}. 
Despite working at fast speeds, the Tiny Video Networks, are outperforming or rivaling the strongest state-of-the-art methods, 
which are tens to hundreds of times slower.


TVNs can find wide application in real-time video understanding tasks, such as mobile phones or robots. Furthermore, they can be used for tasks which require heavy computational loads, such as server-side video processing, or for tasks that can benefit from time and energy savings. Due to their fast speeds, TVNs can also impact video research:
TVN models will allow researchers to further progress the video understanding field, training and exploring future video architectures at very low cost thus accelerating video understanding research.
Code and models for TVNs are available. \href{https://github.com/google-research/google-research/tree/master/tiny_video_nets}{https://github.com/google-research/google-research/tree/master/tiny$\_$video$\_$nets}.
Trained models are publicly available via TF.Hub at
\href{https://tfhub.dev/google/tiny_video_net/tvn1/1}{https://tfhub.dev/google/tiny$\_$video$\_$net/tvn1/1}.
\href{https://tfhub.dev/google/tiny_video_net/mobile_1/1}{https://tfhub.dev/google/tiny$\_$video$\_$net/mobile$\_$1/1}.

\section{Related Work}

Traditionally, building efficient networks has been an important problem with a wealth of research~\cite{mobilenetv2,luo2017thinet,mobilenetv2,tan2018mnasnet,wofk2019fastdepth,wu2019fbnet,zhang2018shufflenet}. Network architectures have also been designed for specific hardware or mobile devices, e.g.,~\cite{howard2019searching,mobilenetv2,tan2018mnasnet,wofk2019fastdepth,wu2019fbnet,cai2018proxyless,xiong2019resource}, where larger networks are optimized to run at fast speeds on these devices. Furthermore, advances in neural architecture search (NAS)~\cite{zoph2017neural,real2017large,liu2019darts,real2019amoeba,zhu2019eena} demonstrated large gains in recognition accuracy and successful results. 
NAS methods also demonstrated automatically building time-constrained models~\cite{howard2019searching,tan2018mnasnet,pham2018efficient,liu2018progressive}, in some cases with hardware in the loop~\cite{yang2018netadapt,wu2019fbnet}.

Many innovative video understanding methods are proposed, some of which also focusing on efficiency~\cite{Alwassel2018action,carreira2018massively,carreira2017quo,chen2018multi,diba2018spatio,diba2019holistic,fan2019more,feichtenhofer2018slowfast,hara2018can,Hussein2019timeception,3dconv,Korbar2019scsampler,lee2018motion,lin2019tsm,luo2019grouped,qiu2017learning,simonyan2014two,su2016leaving,sun2018optical,tran2014c3d,tran2018closer,wu2019multiagent,wu2019adaframe,xie2018rethinking,yeung2015every,eco}.
Architecture search for videos has been scarce~\cite{piergiovanni2018evolving,ryoo2020assemblenet} and had produced relatively expensive networks. Recent work generates efficient video models~\cite{feichtenhofer2020X3D} by expanding fast 2D models. 
We here create efficient video architectures automatically from first principles which are both highly efficient and accurate, with competitive performance to the best  architecture-searched models~\cite{ryoo2020assemblenet}.







Some works reduce the computation cost of video CNNs. Representation flow~\cite{piergiovanni2018representation}, MFNet \cite{chen2018multi} reduce the computation of optical flow, while CoViAR \cite{wu2018compressed} focus on using compressed videos (e.g., MPEGs) to perform recognition. These works rely on heavy CNNs to obtain strong results.
We here focus on obtaining efficient and compact CNN architecture, which itself can run on top of compressed videos.
Furthermore, online video understanding works have generated fast video processing by reusing computations across frames~\cite{eco,lin2019tsm}. These works are complementary to ours, as the fast standalone video architectures we propose can be further utilized in even more efficient online recognition. 


\section{Tiny Video Networks}

We here describe how to automatically design the Tiny Video Networks so that they can solve efficiently a number of video understanding tasks. The goal is to build a number of video architectures, by starting from random ones and iteratively improving their structure until sufficiently good architectures are obtained. We use evolution as the main optimization algorithm~\cite{goldberg91acomparative,real2019amoeba} which enables the construction of new network structures, can run in parallel, and allows for imposing non-differentiable constraints, such as time and memory limits.

Since the networks need to solve challenging video understanding tasks and at the same time be efficient,
the main idea is to create architectures out of basic network building blocks, but allow the selection of effective {\it combinations} of layers, working either in the spatial or time domains, which are successful at the video understanding task. 


In order to apply evolutionary search for videos, several careful considerations need to be made.
(i) We specifically focus on a set of primitive neural layers. For example, we do not include video-specific elements - such as 3D convolutions, since they are expensive. Instead, elements such as pooling and convolutions are used. The architecture search is tasked with discovering successful sequences of them, which satisfy predefined constraints, e.g. runtime. 
(ii) The network layers can chose to work in either the spatial or the time dimensions.
Importantly, specific {\it combination} of layers, e.g. some working in time followed by spatial ones, 
if successful at the video understanding task, will be selected in future evolved architectures and in the final learned architectures. Such sequences of layers 
enable learning spatio-temporal features, and, as seen in the experiments, can be very efficient. 
(iii) The architecture search can decide which input resolution is most advantageous, combined with how many frames per video snippet are needed and at what sampling rate. This allows for further adaptability of the framework to obtain efficient and accurate architectures. 
(iv) We explore building networks constrained for runtime, and optionally model parameters or memory footprint. 
This is beneficial for the network to find directly models applicable to special hardware e.g. for Mobile.
We further take advantage of the {\it constraints} to eliminate quickly models without evaluating their fitness, if they do not satisfy the constraints. This allows for large diversity of architectures and exploration of many models quickly.
(v) We consider highly efficient network elements: 2D spatial or 1D temporal convolutional layer, 1D pooling,  non-local blocks \cite{wang2018non}, context-gating layers \cite{miech2017CG}, and squeeze-and-excitation layers~\cite{hu2018squeeze}.
Furthermore, we explore various other structures, such as,
skip connections and nonlinearities, since these are also important in optimizing the architectures.
(vi) In order to limit the search space we organize the architectures {\it in blocks}, which can be repeated. Each block can contain one of the available layers. This provides a certain structure to the networks, rather than exploring an unconstrained sequences of random numbers of layers. 



\subsection{Tiny Video Architecture Search}

In order to learn efficient video architectures, we maximize the following equation where the input is the set of parameters defining a neural network architecture.
Let $N_{\theta}$ be the network configuration, which corresponds to a specific architecture, and
$\theta$ denote the learnable weights of the network ($|\theta|$ is the number of weights in the network). Let $P$ be a hyperparameter controlling the maximum parameter size of the network, if such a constraint is desired. 
We denote by $\mathcal{R}(N_\theta)$ the function which computes the runtime of the network on a device, given the network $N$ with its weight values $\theta$, and by $R$ the maximum desired computational runtime. We then optimize:
\begin{equation}
\label{eq:main}
\begin{aligned}
& \underset{N_\theta}{\text{maximize}}
& \mathcal{F}(N_\theta) \\
& \text{subject to}
&  \mathcal{R}(N_\theta) < R \\
&&|\theta| < P,
\end{aligned}
\end{equation}
where $\mathcal{F}$ is the fitness function, which measures the accuracy of the trained model on the validation set of a dataset. We optimize Eq. \ref{eq:main} by evolutionary search~\cite{goldberg91acomparative}, starting from random architectures and searching for new network structures, which improve the fitness score, and satisfy the constraints. 
The search is done by measuring runtime on a regular desktop CPU. The fitness function $\mathcal{F}$ we use is the sum of top 1 and top 5 accuracy per model. Optionally, a constraint on memory usage can be imposed similar to parameter size.




We use the tournament selection evolutionary algorithm with discrete mutation operators~\cite{goldberg91acomparative}. 
Since the search space is large, we begin by generating a pool of 200 random networks. 
For example, a network first randomly selects a fixed number of blocks for the network, e.g., as a uniform random variable between 1 and 8, the input image resolution, the number of frames to be sampled from the video snippet, the frame rate, and other parameters. Then, per each block, we randomly sample a sequence of layers. They are selected randomly from a potential set of components such as 2D spatial or 1D temporal convolution layers, pooling, context-gating layers, etc.  A residual connection at the end of a block can also be (randomly) enabled (please see Section~\ref{sec:search_space} for description of the full search space and parameter value ranges).

After evaluating these initial networks, we randomly choose 50 of them and take the top performing network as a `parent'. This is typically done in parallel, by 10 workers. 
We then apply a discrete `mutation' operation to this network by randomly changing one part of the network according to the search space. For example, randomly changing either the input resolution, the number of blocks, or a layer within a block. This allows for the `pool' of architectures to stay diverse and at the same time to evolve to a set of potentially better architectures at each round.

The models are partially trained to 10,000 iterations in order to evaluate their performance.
We ran the evolution until a saturation point of about 500 rounds. 
The average training time per model is about 1.5 hours, but actual training times may vary significantly, as well.
Networks which do not satisfy the runtime or other constraints are quickly discarded, speeding up the search and contributing to the overall fast evolution process. Taking advantage of parallel training,
the full search is done within a day. 


Evolutionary search provides several advantages.
An evolutionary algorithm easily allows targeting  different types of devices within the search. It effectively explores the irregular search space with a non-differentiable objective function, e.g. adding constraints on the number of parameters or the model memory footprint, which are important for mobile applications. 
Evolution
allows for parallelizing the search which significantly speeds it up. 
Since each architecture considered in the search is extremely efficient to begin with, the search itself is not as computationally intensive as other architecture search methods, as only sufficiently fast networks are evaluated in the search, and the others are discarded. All these considerations are important for videos, as they are computationally very demanding.

\subsection{Search space}
\label{sec:search_space}
In order to generate a random network to start the evolution, one can sample from each of the components of the search space. 
For example, a network first randomly selects the input resolution which can be between ($32\times 32$ to $320\times 320$) with a step size of 32. Then it will randomly pick the number of frames (1-32), and framerate - 1fps to 25fps (this is also referred to as `stride', i.e. number of frames to skip; the stride is selected by uniform random sampling but depends on the number of frames already selected). Then it selects a fixed number of blocks as a uniform random variable between 1 and 8, and number of `repeats' per block (up to 8). Then, per each block, we randomly sample a sequence of layers. They are selected randomly from a potential set of components such as 2D spatial or 1D temporal convolutional layer, 1D pooling,  non-local blocks \cite{wang2018non}, context-gating layers \cite{miech2017CG}, and squeeze-and-excitation layers~\cite{hu2018squeeze}.  A residual connection at the end of a block can also be (randomly) enabled. Blocks are used for simplicity only and are not required. Their use reduces the search space somewhat, as the structure imposed by the blocks, eliminates some combinations. We found that this is still a very effective and little-constraint search space.  

For each of these layers, a specific set of parameters are also sampled, in order to fully form a computational layer.
For non-local layers, we search for the bottleneck size (between 4 and 1024). 
We search for the squeeze ratio for the squeeze-and-excitation layers (a real-valued number between 0 and 1). The convolutional layers can have a variety of kernel sizes (from 1 to 8), strides (from 1 to 8), number of filters (from 32 to 2048) and types (e.g., standard convolution, depthwise convolution, average pooling or max pooling). 
Additionally, a layer can optionally pick an activation function, a ReLu (or a swish~\cite{howard2019searching} for the Mobile-friendly models).
The final block is followed by a fixed standard block of global average pooling, a dropout layer (0.5 fixed dropout rate), and a fully-connected layer which outputs the number of classes required for classification.

The mutation operations, again follow the search space described above. 
A single mutation is allowed per network, and for example it can be at the high-level, i.e. changes to the resolution, number of frames, or number of blocks (here adding or removing a block from an existing net is allowed, so as to preserve all other functionalities of the net). Mutation can also be at the level of some parameters - e.g. changing the number of repeats per block, type of layer in a randomly sampled block and layer, or changing the number of filters or other parameter of a specific layer.

Since we are working with videos, exploring all of these potential architectures leads to a very large search space. Each block has $\sim 2^{34}$ possible configurations. When including the input resolution and up to 8 blocks in the network, the search space has a size of $\sim 2^{45}$ or about $\sim 10^{13}$. Thus, without automated search (e.g., if we do a brute-force grid search or random search), finding good architectures in this space is prohibitive. 
We here use such a large and unconstrained search space to find the optimal use of temporal and spatial information within a specific computational budget. 
In our experiments, we observe that many of the architectures in this space give poor classification results, diverge during training, or exhibit other unstable behaviors.
Other networks are also quickly discarded as they do not meet the timing criterion, giving space to new viable architectures, which helps the search, and also quickly constrains the search space to already good architectures.
We find that the evolutionary search is not very sensitive to its hyper parameters, so long as sufficient population size and rounds are allowed: naturally each new evolution process generates a new set of architectures, and although different, the best ones perform very similarly.


By default, we run the search, optimizing for runtime on CPU. We can achieve better GPU runtimes, or FLOPs, if optimized for them. Optionally, we constrain the number of parameters or the memory footprint. Surprisingly, we find the search may generate networks with large number of parameters which still run very efficiently, compared to other similar-sized models, by taking advantage of wide parallel layers rather than deep sequential layers. The search space can be modified to include new elements, as we do for Mobile-specific TVNs (Sec.~\ref{tab:mobile-search}).
 An expanded search space, more extensive search, e.g., larger pool, more mutations, and optimization for a different device, can be explored. 



\section{Experiments}


We conduct experiments on four challenging video datasets: Moments-in-Time, Charades, HMDB and MLB. 
By placing different constraints on the search space, we automatically generate TVNs of various capacities and runtimes (Figure~\ref{fig:intro}).
TVNs are evolved on different datasets to capture various aspects of the specific video dataset scenarios.
We test their usability across datasets:
models evolved on one dataset are cross-evaluated on all other datasets.

\vspace{-0.1cm}
\begin{table}
\begin{tabular}{llccccc}
\toprule
Model & Evolved on  & Runtime &Image & $f$ & $s$ \\
  &  &CPU/GPU  &resol. & & \\
\midrule
TVN-1 &MiT & 37 / 10   &224x224  &2  &4 \\
TVN-2 &MLB & 65 / 13   &256x256  &2  &7 \\
TVN-3 &Charades & 85 / 16 &160x160  &8  &2 \\
TVN-4 &MiT & 402 / 19  &128x128  &8  &4  \\
TVN-5 &MiT & 86 / 16  &160x160   &16  &4 \\
TVN-6 &MiT & 142 / 18   &160x160  &32  &4 \\
\bottomrule
\end{tabular}
\label{tab:tvns}
\caption{TVN models. Runtime is in milliseconds (ms).}
\vspace{-0.6cm}
\end{table}

\subsection{Found TVN Models} 
\label{sec:found}
Our method finds unique, yet simple and elegant architectures which are multiple times more efficient than other networks (Figure~\ref{fig:intro}). 
Figure~\ref{fig:teaser} shows example TVNs; they
 feature deeper architectures of fast layers, and an unusual sequence of layers
(see sup. materials for more).
Interestingly, non-local layer~\cite{wang2018non} is rarely preferred in TVNs. This suggests that it is more cost-efficient to spend computation on deeper and/or wider networks. Layers, such as context gating, which are significantly cheaper, are commonly used. 

We describe the found TVNs (Table~\ref{tab:tvns}): 
TVN-1 is the fastest model found. It was evolved on the MiT dataset by constraining the search space to include models only running in less than 50ms on CPU; it runs at 37ms on CPU and 10ms on GPU. 
TVN-2 is evolved on the MLB dataset by limiting the search space to 100ms and 12 million parameters; it runs at 65ms. 
TVN-3 was evolved on Charades by limiting the search space to 100ms as well, but no constraint on the number of parameters; it runs at 85 ms on CPU. 
TVN-4 is a slower model, which we created trying to understand what performance we can get allowing the model to take more time for inference. It is found by allowing networks up to 1200ms and 30 million parameters (a max computation cost roughly comparable to I3D). It runs at 402ms on CPU and is evolved on Moments-in-Time. 
TVN-5 is found by assigning limit on runtime and memory usage: up to 100ms and 2.5GBytes. Because of the additional memory contraint it manages to create a deeper model which is still very fast and is quite accurate across datasets. It runs at 86ms on CPU and 16 ms on GPU and is evolved on MiT. TVN-6 is the same as TVN-5 but we set 32 frames as input. Thus this model will be using the same number of frames other models in prior art are using. TVN-6 runs at 142ms on CPU and 18ms on GPU.
As seen, both of these models, despite using a large number of input frames are very efficient (Table~\ref{tab:tvns}), which can be attributed to also limiting the memory footprint.
TVNs also learn specific runtime image resolutions, number of frames $f$, and sampling stride $s$. 




\begin{table}
\centering
\begin{tabular}{lccc}
\toprule
Method & Runtime   & GFlops & mAP \\
       &  \small{CPU/GPU} &  &  \\

\midrule
Asyn-TF, VGG16 \cite{sigurdsson2016asynchronous} &-   &-  & 22.4 \\ 
I3D~\cite{carreira2017quo}   &-    & 216  & 32.9 \\
Nonlocal, R101\cite{wang2018non} &-   & 544 $\times$ 30 &  37.5 \\ %

SlowFast, two-str. \cite{feichtenhofer2018slowfast}   & 3594/135 & 213 $\times$ 30 & 42.1\\
SlowFastNL, two-str.  \cite{feichtenhofer2018slowfast} & 4354/152 & 234  $\times$ 30 &  45.2 \\

X3D-XL (Kin400 pretr)~\cite{feichtenhofer2020X3D}  &-   &48.6 $\times$ 30 &43.4 \\
X3D-XL (Kin600 pretr)~\cite{feichtenhofer2020X3D}  &-     &48.6 $\times$ 30 &\it{\textbf{47.1}} \\
\midrule
TVN-1 (from scratch) & 37/10 & 13  & 40.4 \\  
TVN-2 (from scratch) & 65/13 & 17 & \textbf{47.4}  \\ 
TVN-3 (from scratch) &  85/16 & 69 &\textbf{52.0} \\ 
TVN-4 (from scratch) & 402/19 & 106 &\textbf{53.8}   \\ 
TVN-5 (from scratch) & 86/16 &52 & \textbf{52.4} \\   
TVN-6 (from scratch) & 142/18 &93 & \textbf{52.8} \\   
\midrule
TVN-1 (MiT pretr) & 37/10 & 13  &42.1 \\
TVN-2 (MiT pretr) & 65/13 & 17 &\textbf{48.3}   \\
TVN-3 (MiT pretr) &  85/16 & 69 &\textbf{53.2}  \\
TVN-4 (MiT pretr) & 402/19 & 106 &\textbf{53.9}   \\
TVN-5 (MiT pretr) & 86/16 &52 &\textbf{54.2}  \\   
TVN-6 (MiT pretr) & 142/18 &93 & \textbf{54.6} \\   
\bottomrule
\end{tabular}
\caption{Comparison to the state-of-the-art on Charades. We report the best TVN models in bold, and the best prior work models as bolded italics.
Many TVNs, even without pretraining, outperform the SOTA which uses strong Kinetics pre-training. TVNs also outperform or rival the strongest video models which use additionally flow: AssembleNet~\cite{ryoo2020assemblenet} with flow achieves 47.0 from scratch, and 53.0 with MiT pretraining; AssembleNet++~\cite{ryoo2020assemblenetplus} with flow achieves 54.98. 
TVNs are also orders of magnitude faster.
}
\label{tab:charades-sota}
\vspace{-0.5cm}
\end{table}

\subsection{Experimental setup}

\textbf{Datasets.} We conduct experiments on four well-established public video datasets, representing various challenges for video understanding: 

\textbf{Moments-in-time (MiT) \cite{monfort2018moments}} is a large-scale dataset with 800k training examples and 33900 validation examples. 
It is a challenging dataset with a  large number of classes (339), some of which may be abstract concepts. 

\textbf{HMDB \cite{kuehne2011hmdb}} is a dataset of about 5000 training and about 1500 test examples for 51 different classes. It is a popular dataset used my many prior approaches, although it is relatively small (only about 100 videos per activity).

\textbf{Major League Baseball (MLB) \cite{mlbyoutube2018}} has 4290 videos for 8 different baseball activities. It is a multi-class, multi-label dataset (i.e. more than one label is correct). Unlike the previous two datasets, where fewer frames may provide reasonable performance, MLB contains longer videos and requires understanding temporal information as the actions are fine-grained and occur in the same scene, e.g., `bunt' and `swing' activities are very similar. We use the segmented video setting, as in~\cite{piergiovanni2017learning}.

\textbf{Charades \cite{sigurdsson2016hollywood}}  is also a multi-class multi-label dataset of about 8000 training and 1686 validation videos of 157 different in-home activities. Charades contains long, continuous videos (30 seconds on average) with multiple activities which can be occurring or co-occurring. 
It is also a very challenging dataset due to the large number of actions, the  multi-class multi-label setting, and the small number of examples per class.

Each of these datasets has a varying number of frames per video: MiT has about 37, MLB has about 80, HMDB - about 40 and Charades has about 350-400 frames per video.
During evaluation, the video is sampled every $s$-th frame and the others are discarded. Out of the remaining frames, $f$ frames are picked at random which are the inputs for evaluation, which is standard for video processing. 
Note that TVNs select very low values for these, Table~\ref{tab:tvns}.

We use the established evaluation protocols for all datasets. We report runtime (on CPU and GPU), FLOPs and accuracy or mAP, in the context of state-of-the-art (SOTA) models. We measure runtime on an Intel Xeon CPU running at 2.9GHz and a single V100 GPU. We follow the specified network and inputs for each model: our baselines use 32 frames, as in (2+1)D ResNet \cite{tran2018closer}; I3D \cite{carreira2017quo} and S3D \cite{xie2018rethinking} use 64 frames. 
In contrast, each TVN has learned a number of frames $f$ to sample which is typically small, e.g. $f$=2 for TVN-1 which is evolved on MiT, whereas TVN-3, evolved on Charades, has $f$=8. This means that TVNs see much lower number of frames than other models at inference.  



We report results using RGB as inputs, since flow computation itself is quite expensive and is not suitable for real-time video understanding.  Fast flow representations (\cite{lee2018motion}) still need about 100ms on GPUs, compared to our entire TVNs running in less than $20$ms on GPUs. 



\begin{table}
\centering
\begin{tabular}{lcccc}
\toprule
Method & Runtime & Runtime   & GFlops  & Acc. \\
       &  \small{CPU(ms)} &\small{GPU(ms)}  &  &(\%) \\
\midrule
ResNet-18 & 2120 & 105 & 38 & 21.1 \\
ResNet-34 & 2256 & 110 & 50 &  24.2 \\
ResNet-50 & 3022 & 125 & 124 &  28.1 \\
ResNet-101 & 3750 & 140 & 245 &  30.2 \\
TSN~\cite{wang2016temporal}  &- &- &-  &24.1  \\
2DResNet50~\cite{monfort2018moments}  & -  & - &-  & 27.1 \\

bLVNet-TAM~\cite{fan2019more}  &- &- &- &\it{\textbf{31.4}} \\

\hline

TVN-1  & 37 & 10 & 13   &23.1 \\
TVN-2  & 65 & 13 & 17  & 24.2 \\
TVN-3  & 85 & 16  &69  & 25.9 \\  
TVN-4  & 402 & 19 & 106 & 28.0  \\  
TVN-5  & 86 & 16 & 52 & \textbf{29.8} \\
TVN-6  & 142 & 18 & 93  & \textbf{30.7} \\
\bottomrule
\end{tabular}
\caption{Results on the MiT dataset comparing different Tiny Networks to baselines and state-of-the-art (which are all RGB-only). 
TVN models perform competitively and are also much faster.
No runtime was reported in prior work.}
\label{tab:mit-1}
\vspace{-0.3cm}
\end{table}

\begin{figure}
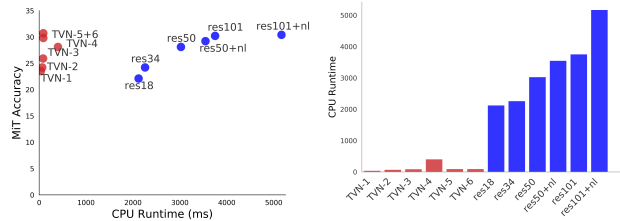

    \centering
     \includegraphics[width=0.49\linewidth]{figures/tvn-plot-new-v2.pdf}
     \includegraphics[width=0.49\linewidth]{figures/tvn-cpu-v2.pdf}
    \caption{
    Runtime vs. model accuracy of TVNs compared to ResNet video recognition baselines (left) and CPU runtime (right).
    TVNs reside in a very good portion of the runtime-vs-accuracy space, of very accurate and fast models. 
    They are orders of magnitude faster than contemporary video models, e.g. ResNet-101.
    }
    \label{fig:performance_acc}
    \vspace{-0.3cm}
\end{figure}

\subsection{Comparison to the state-of-the-art}


 

\textbf{Charades:}
Table~\ref{tab:charades-sota} shows the performance of the Tiny Video Networks evaluated on the Charades dataset.
TVNs obtain impressive results over SOTA. 
Firstly, TVN-2, TVN-3, TVN-4, TVN-5, TVN-6, without using any pre-training, already outperform the best state-of-the-art models which used powerful Kinetics pre-training. 
Among the SOTA are strong models with much more compute e.g. an architecture searched X3D~\cite{feichtenhofer2020X3D} and the two-stream SlowFast models~\cite{feichtenhofer2018slowfast}. They are 40-100 times slower and use 15-75 times more GFLOPs (e.g. for TVN-6). 
For this dataset, we also fine-tune the models trained on MiT, as is customarily done in the literature\footnote{Our results are also disadvantaged as we pre-train on MiT. Kinetics pretraining provides better performance than MiT~\cite{ryoo2020assemblenet}. For reasons not dependent on us, we are unable to use the Kinetics dataset.}, and obtain even better results.
Curiously, TVNs rival performance of the most powerful video models to date, AssembleNet and AssembleNet++~\cite{ryoo2020assemblenet,ryoo2020assemblenetplus}, which use additionally optical flow and connectivity learning (see caption).
We note that achieving such performance at a fraction of the speed is an impressive result.



\begin{table}
    \begin{minipage}{.5\textwidth}
  
\begin{tabular}{lccc}
\toprule
Method & Runtime & Runtime & mAP \\
       &  (CPU) & (GPU) &   \\
\midrule
InceptionV3 & - & - & 47.9 \\
I3D~\cite{mlbyoutube2018} & 1865ms$^*$ & -  & 48.3 \\
I3D+sub-events~\cite{mlbyoutube2018} &- & - & \it{\textbf{55.5}} \\
\midrule
TVN-1  & 37ms & 10ms  & 44.2 \\
TVN-2  &  65ms & 13ms  & 48.2 \\
TVN-3  & 85ms & 16ms   & 46.5 \\
TVN-4  & 402ms & 19ms &  52.3 \\
TVN-5  & 72ms & 16ms  &\textbf{55.3} \\
TVN-6  & 142ms & 18ms  & \textbf{56.4} \\
\bottomrule
\end{tabular}
 \caption{Comparison to the state-of-the-art results on MLB. No prior runtimes are available. $^*$Our measurement of runtime. }  
    \label{tab:mlb}
 \end{minipage}\hspace{14mm}%
\begin{minipage}{.4\textwidth}
    
\begin{tabular}{lcc}
\toprule
Method & GFlops  & Accuracy (\%) \\
\midrule
CoViAR~\cite{wu2018compressed} &-   & 59.1 \\
I3D~\cite{carreira2017quo}   &300      & 74.8 \\
S3D-G~\cite{xie2018rethinking}     &-  & \it{\textbf{75.9}} \\
ECO (online)~\cite{eco}  &64  &72.4 \\
TSM (online)~\cite{lin2019tsm}  &65  &73.5 \\
\midrule
TVN-1  & 13   & 72.1 \\
TVN-2  &  17   & 73.5 \\
TVN-3 & 69 & 71.8  \\
TVN-4  & 106  &  74.7 \\
TVN-5  & 52    & 73.4 \\
TVN-6  & 93    & \textbf{75.5} \\
\bottomrule
\end{tabular}
\caption{Performance on HMDB. }    
    \label{tab:hmdb-sota}
 \end{minipage}
 \vspace{-0.3cm}
\end{table}

\textbf{Moments-in-Time:}
Table~\ref{tab:mit-1}
shows the main results of TVNs on the MiT dataset. 
The TVNs perform competitively or better than previous state-of-the-art methods, while running extremely fast.
Since prior work did not report runtime, we also compare to strong  baselines, (2+1)D ResNets \cite{tran2018closer,wang2018non} (Figure~\ref{fig:performance_acc}).
 For example, TVN-1 and TVN-2, both outperform  ResNet-18 and are 57 and 33 times faster, respectively; they are at the same performance as ResNet-34, while being 61 and 35 times faster. 
TVN-6 outperforms ResNet-101, being 26 times faster.
We also see that models that are evolved on the MiT dataset itself tend to perform better on it than models of the same capacity but evolved on other datasets (e.g. TVN-5 vs TVN-3), which is expected as videos vary in their characteristics across datasets.

In terms of compute, TVN models perform very well even compared to prior {\it online} video models. For example, ECO and TSM~\cite{eco,lin2019tsm} have at least three times the GFLOPs (64 and 65, respectively) than TVN-1 and TVN-2 (13 and 17, respectively), larger GFLOPs than TVN-5 and comparable to TVN-3. Given that TVNs are standalone models, they can further benefit from online inference.


\textbf{MLB:}
Table~\ref{tab:mlb} shows the performance of TVNs on the MLB dataset, which targets fine-grained actions.
We see here too that TVNs outperform SOTA, which is impressive for their runtimes. 
For example, TVN-5 and TVN-6 outperform I3D while being 25 and 13 times faster, respectively. TVN-6 outperforms~\cite{mlbyoutube2018}, being at least 13 times faster.

\textbf{HMDB:}
Table~\ref{tab:hmdb-sota} shows TVN performance on HMDB. Here we report the `averaged over 3 splits', per the standard evaluation protocol. We did not evolve models on HMDB as it is a very small dataset. 
TVNs are comparable to~\cite{xie2018rethinking} on HMDB, and outperform online models ECO and TSM~\cite{eco,lin2019tsm}, where ECO further decreases its performance to 68.5 when using 16 frames, and to 61.7 when using 4 (TVN-1 achieves 72.1 with 2 frames, and TVN-4 is at 74.7 with 8).

While in general, we observe that the model evolved on its own dataset performs better than others, larger-capacity models, e.g. here TVN-4, TVN-5, TVN-6 perform best. 
We note that the runtimes per-model do not change per-dataset, since the model itself has selected the input resolution and the frame rate, and naturally that may be a sub-optimal choice for other datasets.



\textbf{Number of parameters.}
We note that there is not a direct relationship between parameters/flops and runtime due to hardware differences, and the configuration of layers which TVN search is designed to optimize, has a larger effect on runtime.

\begin{wrapfigure}  {r}{0.425\linewidth}
    \centering
    \includegraphics[width=0.9\linewidth]{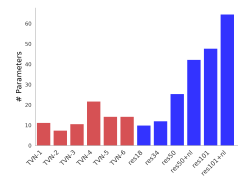}
    \caption{Number of model parameters (in Millions) of TVNs (in red) vs ResNets. }
    \label{fig:param}
\end{wrapfigure}

Figure~\ref{fig:param} shows the number of parameters of the TVN models.  For a side-by-side comparison with their corresponding runtimes please see Figure~\ref{fig:performance_acc}). 
Naturally, smaller models have fewer parameters.  
However, some of the faster TVNs have more parameters, e.g., TVN-1 has more parameters than TVN-2, ResNet-18 and Resnet-34, but is much faster. 
TVN-3 has more parameters than ResNet-18, ResNet-34, but is faster and more accurate.
TVN-4, TVN-5, TVN-6 are both more accurate and faster than other powerful ResNet models, despite having several times fewer parameters. 
This is due to the ability of TVN architecture search to build architectures taking advantage of specific characteristics of the hardware, i.e. TVNs are 
evolved to use the computational resources more efficiently. 
 


\section{Tiny Video Models for Mobile Deployment}

\subsection{Mobile-friendly Tiny Video Networks}

We further make a modification to our search space to include mobile-friendly components, such as inverted residual layers and the hard swish activation function, similar to MobileNet~\cite{mobilenetv2,howard2019searching}, applied both in space and time dimensions.
Since search is done with the same constraints to be comparable to the original TVNs, i.e., runtime within 100ms on CPU, there is no guarantee that Mobile-only components will be selected. Still the search is able to uncover more interesting TVNs. Table~\ref{tab:mobile-search} shows two selected mobile models, named TVN-M-1 and TVN-M-2. They are comparable to TVN-1, but achieve 23\% fewer Flops and have almost twice fewer parameters, which are both important for mobile, at a small reduction in accuracy. These models satisfy all requirements for production Mobile deployment.
Furthermore, we modify the original TVN-1 by substituting all ReLu activations with the hard-swish~\cite{howard2019searching}, we find an improvement in accuracy of 1.7\% with only negligible 2ms loss in runtime, confirming its usefulness.

\begin{wrapfigure}  {r}{0.425\linewidth}
   \vspace{-0.5cm}
    \includegraphics[width=1.0\linewidth]{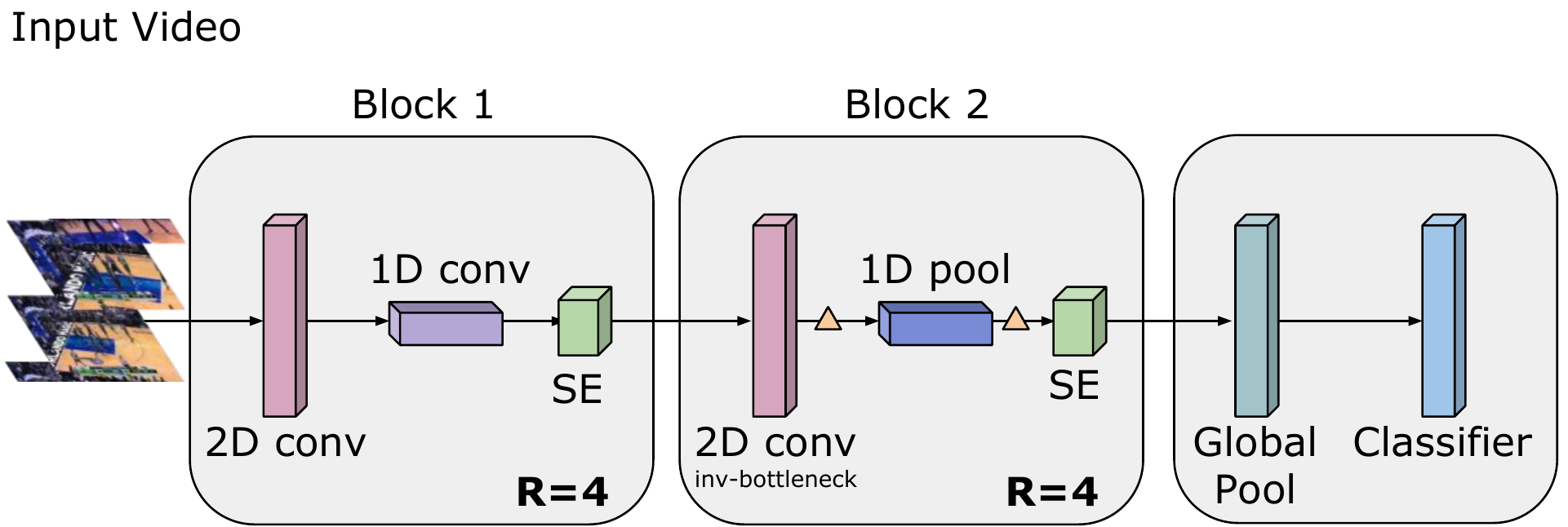}
    \caption{Illustration of TVN-Mobile. The triangle is a hard-swish activation. 
    }
    \label{fig:tvn-mobile}
    \vspace{-0.1cm}
\end{wrapfigure}


\begin{table}
\centering
\begin{tabular}{lcccc}
\toprule
Method & Runtime  & Params. &GFlops & Accuracy \\
       &  (CPU) &  &  \\
\midrule
TVN-1 & 37ms & 11.1M &13.0 & 23.1\% \\
TVN-1 + swish & 39ms & 11.1M &13.0 & 24.8\% \\
\midrule
TVN-M-1 & 43ms & 5.6M &10.0 &21.95\% \\
TVN-M-2  & 75ms & 5.4M &10.1 &  21.96\% \\
\bottomrule
\end{tabular}
\caption{TVN models obtained when expanding the search with components used in MobileNet. MiT.}
\label{tab:mobile-search}
\vspace{-0.3cm}
\end{table}

Figure~\ref{fig:tvn-mobile} visualizes one of the networks built with mobile components (here, TVN-M-1, on MiT). The inv-bottleneck is the inverse bottleneck layer used by MobileNet \cite{mobilenetv2}, which generally saves parameters without harming performance. 

\begin{table}[]
    \centering
    \begin{tabular}{c|cc|cc}
    \toprule
        \# Frames & MobileNet  & MobileNet   & TVN   & TVN     \\
    
          & Runtime & Accuracy & Runtime &  Accuracy   \\
    \midrule
      1 Frame	& 42ms	& 18.8  &32ms  &20.2 \\
      2 Frame	& 58ms	& 19.3  &37ms   & 23.1 \\
      8 Frame	& 280ms & 	20.8  &85ms  &25.9\\
    \bottomrule
    \end{tabular}
    \caption{MobileNet results compared to Tiny Video Nets. As seen, TVNs outperform them in both faster rutimes and higher accuracies, when applied to videos. MiT dataset.}
    \label{tab:mobilenet}
    \vspace{-0.3cm}
\end{table}

\subsection{Comparison to MobileNet models.}

MobileNetV3~\cite{howard2019searching} models are among the fastest single-image models to date, and are also obtained via a neural architecture search. It is interesting to see how such models can perform when adapted to videos.
We compare our Tiny Video models to a MobileNetV3-equivalent ones, by applying  MobileNet per frame, with a max pooling before the final fully connected layer, and training this video-adapted model. We report the performance of TVNs which have 2 and 8 frames, and of a single-frame TVN model, which we evolved on MiT, with accuracy of 20.2\%, 32ms runtime, 8 GFlops, at 224x224 resolution. Table~\ref{tab:mobilenet} shows that TVNs are advantageous in both accuracy and speed, especially notable are the significant improvements in both directions for larger number of frames.  

\section{Ablation results}


We conduct a number of ablation experiments.

\textbf{Exploring a range of number of frames.}
One key advantage of TVNs is the opportunity to select the number of frames and how to sample them. For some datasets, e.g. MiT, the network prefers to use very few frames 2-4 to reduce the computation cost. This is expected, given that many activities there are scene-based (e.g. swimming, baseball), so a single frame is often enough to discriminate them (see a 1-frame model in Table~\ref{tab:mobilenet}, top row, performing reasonably on MiT, even though using a single input frame).

To determine the effect of temporal information on performance, we increase the number of inputs frames used by TVN-1 from 2 to 8 and 16, and re-trained these models on MiT, providing more input information to the model (Table \ref{tab:frames}). 
We find that increasing the number of frames for TVN-1 on MiT does not lead to significant performance increase, while the runtime increases a lot. 
At the same time our evolved TVN-5, which also has 16 frames, is 2.3 times faster and is by 6.3\% more accurate on MiT (see Table~\ref{tab:mit-1}).

\begin{table}
\centering
\begin{tabular}{lccc}
\toprule
Method & Runtime  & Runtime & Accuracy \\
       &  (CPU) & (GPU) &  \\
\midrule
TVN-1 (2 frames) & 37ms & 10ms & 23.1\% \\
TVN-1 (8 frames) &  140ms & 28ms & 23.4\% \\
TVN-1 (16 frames) &  200ms & 45ms  & 23.5\%  \\
\bottomrule
\end{tabular}
\caption{Increasing the number of frames for TVN-1 from 2 to 16 on MiT. Just adding more frames as input is not greatly beneficial.}
\label{tab:frames}
\vspace{-0.3cm}
\end{table}



\textbf{Scaling Up the TVNs.}
We further demonstrate the performance of the models by scaling up the found TVNs in image resolution, width and depth (see the supp. material for a detailed experiment).
We also apply an EfficientNet-style scaling \cite{tan2019efficientnet}, i.e. scaling all dimensions (input resolution, width and depth). 
Table \ref{tab:scale-up-mit} reports the best scaling results from both. The EfficientNet-scaled model, denoted TVN-1 EN, achieves higher performance than other scaling versions, and outperforms ResNet-50, being 10x faster.
This is an interesting result, as an easy way of generating not-so-small but still efficient model.
Still, as seen, specifically evolved TVNs are both more accurate and faster than scaled ones (e.g. TVN-5, TVN-6, see Table~\ref{tab:mit-1}).


\begin{table}
\centering
\begin{tabular}{lccc}
\toprule
Method & Runtime  & Runtime & Accuracy \\
       &  (CPU) & (GPU) &  \\
\midrule
(2+1)D ResNet-50 & 3022ms & 125ms & 28.1\% \\
\midrule
TVN-1 (4x wide) &  275ms & 60ms  & 24.2\%  \\
TVN-1 EN &  305ms & 92ms  &  28.2\% \\
\midrule
TVN-5 &  72ms & 16ms  &  29.8\% \\
\bottomrule
\end{tabular}
\caption{Scaling up our tiniest model (TVN-1) on MiT can easily reach ResNet performance for faster speeds.
}
\label{tab:scale-up-mit}
\end{table}

\begin{figure}
    \centering
    \includegraphics[width=0.43\linewidth]{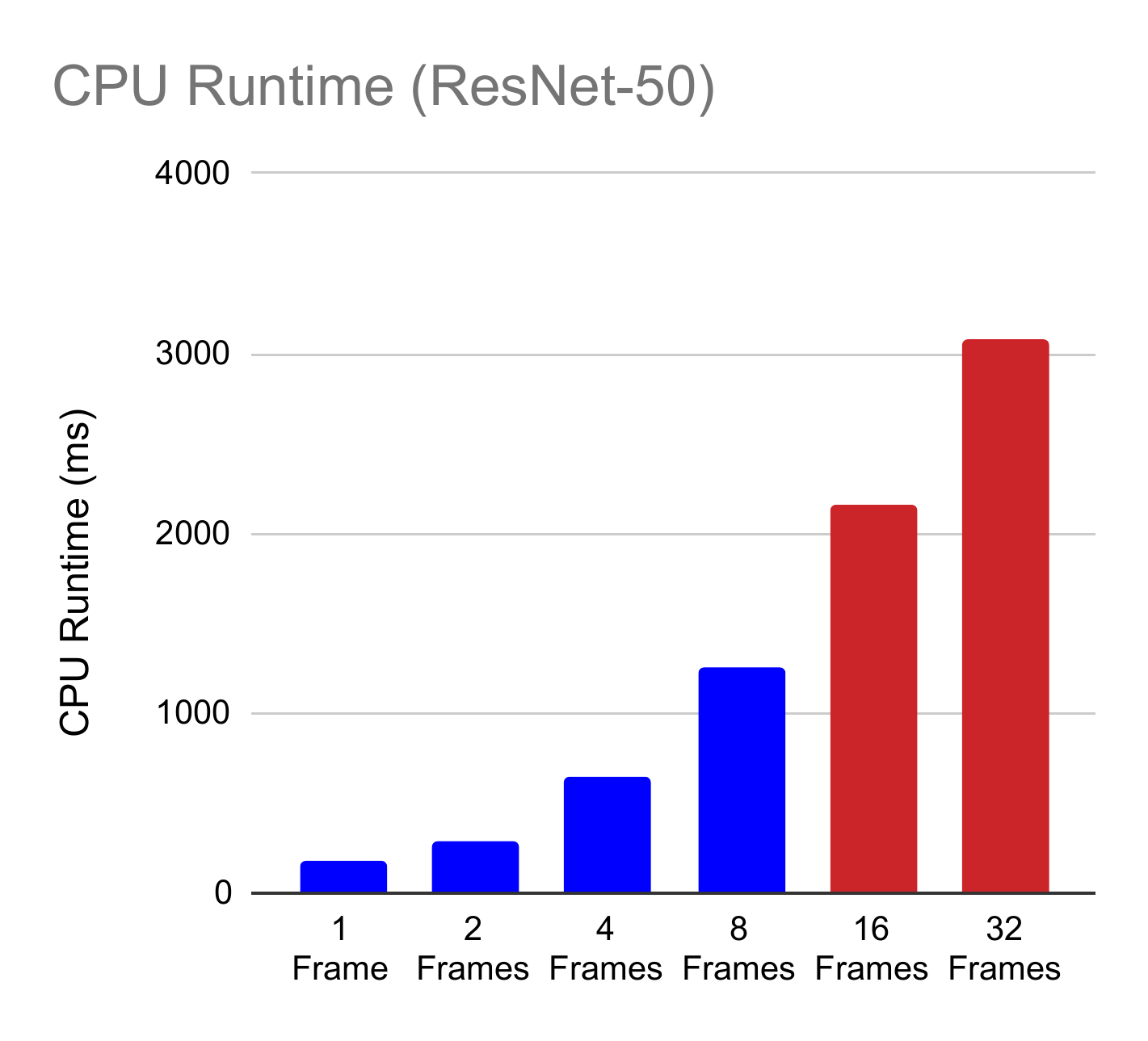}
    \includegraphics[width=0.43\linewidth]{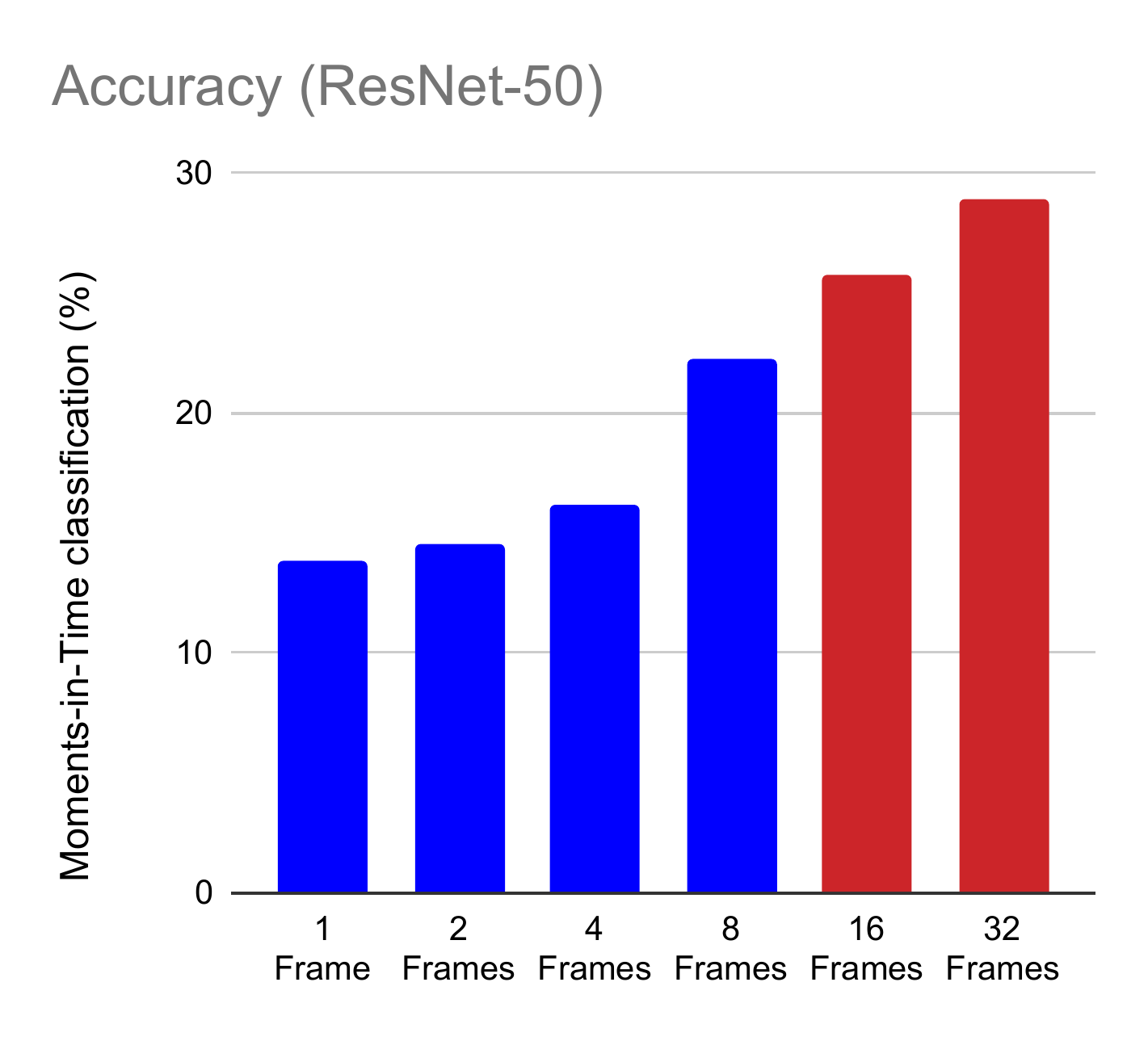}
    \caption{Comparing different number of frames using a standard (2+1)D ResNet-50.
    The last two bars (in red) are for models with accuracy higher than the accuracy of TVN-1. As seen, even using 1-frame (2+1)D ResNet-50 has 185ms runtime on CPU, far slower than a TVN, with much poorer accuracy. More accurate ResNet-50 are much slower with runtime of more than 2000ms on CPU.}
    \label{fig:resnets}
\vspace{-0.5cm}
\end{figure}

\textbf{Comparison to (2+1)D ResNet-50.}
We further compare TVNs to standard ResNet-50 models for the same number of frames. 
In Figure~\ref{fig:resnets}, we show performance of (2+1)D ResNet-50 with varying number of input frames. Even using 1-frame, a ResNet-50 has 185ms runtime on a CPU, far slower than a TVN. Further, in terms of accuracy, TVN-1 outperforms ResNet-50 until 16 frames are used. 

\section{Conclusion}
We propose the Tiny Video Networks, which are automatically designed efficient video architectures. Despite working at unprecedented speeds, 
they accomplish strong results or outperform SOTA on four video datasets.
There are two major impact opportunities of this work:
(i) TVNs can be deployed in real-life applications, e.g., robotics systems or mobile phones. 
Our mobile-friendly TVNs satisfy runtime and parameter constraints for mobile deployment. 
(ii) Our approach provides for easier and faster exploration of video architectures at low cost, which
will allow other researchers to advance the video understanding field. 
We have prepared code and models for open-sourcing.

\clearpage

\appendix


{\small
\bibliographystyle{ieee_fullname}
\bibliography{bib}
}

\end{document}